# On Choosing Training and Testing Data for Supervised Algorithms in Ground Penetrating Radar Data for Buried Threat Detection

Daniël Reichman, Leslie M. Collins, and Jordan M. Malof
Department of Electrical & Computer Engineering, Duke University, Durham, NC 27708

*Abstract*— Ground penetrating radar (GPR) is one of the most popular and successful sensing modalities that has been investigated for landmine and subsurface threat detection. Many of the detection algorithms applied to this task are supervised and therefore require labeled examples of target and non-target data for training. Training data most often consists of 2-dimensional images (or patches) of GPR data, from which features are extracted, and provided to the classifier during training and testing. Identifying desirable training and testing locations to extract patches, which we term "keypoints", is well established in the literature. In contrast however, a large variety of strategies have been proposed regarding keypoint utilization (e.g., how many of the identified keypoints should be used at targets, or non-target, locations). Given the variety keypoint utilization strategies that are available, it is very unclear (i) which strategies are best, or (ii) whether the choice of strategy has a large impact on classifier performance. We address these questions by presenting a taxonomy of existing utilization strategies, and then evaluating their effectiveness on a large dataset using many different classifiers and features. We analyze the results and propose a new strategy, called PatchSelect, which outperforms other strategies across all experiments.

*Index Terms*—training, ground penetrating radar, landmine detection

## I. INTRODUCTION

A popular approach for detecting buried threats, such as landmines and other explosive hazards, is the use of remote sensing technologies. One of the most successful modalities for remote sensing of buried threats is the ground penetrating radar (GPR) [1]–[5]. The typical GPR consists of an array of antennas that are directed toward the ground. An individual GPR antenna operates by emitting a radar signal towards the ground and then measuring the energy that is reflected back. The result of this sensing process is a time-series of energy measurements for the given antenna, referred to as an A-scan [6].

In the context of buried threat detection (BTD), GPR sensors collect A-scans at regular spatial intervals as they move across the surface of the ground (e.g., on the front of a vehicle as it drives forward). The resulting A-scans, each collected at a different spatial location, can then be concatenated to form images of the subsurface, termed B-scans [1], [2], [7]. B-scans have one spatial axis, and one temporal axis. The signals returned from buried threats typically exhibit characteristic hyperbolic patterns in the B-scans, which can be leveraged for detection [6], [8]–[10]. Figure 1 shows several examples of B-scans collected over buried threats.

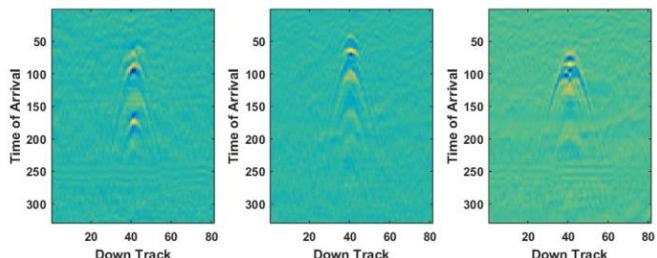

Figure 1: Examples of three B-scans collected over three different buried threats. Each column of this image is the time series of data (i.e., time of arrival, the y-axis) collected at a single spatial location. Consecutive A-scans, for a single antenna, collected as the vehicle travels down the lane (i.e., down track, the x-axis), are shown here as an image. The target signatures in this figure illustrate some of the variability in the response that is possible in GPR data. This includes varying the y-offset of the top of the signature, having several independent regions, and extending over a very different number of time samples.

Although it is possible to manually identify buried threats in GPR data, a great deal of published research has focused on automating this process with computer algorithms that provide a confidence of buried threat presence at each spatial location [4], [7], [8], [11]–[16]. Proposed algorithms have employed a variety of techniques from statistics [17], [18], computer vision [6], [19], [20], and machine learning [7], [21], [22]. The most successful approach to date involves the use of supervised learning techniques [12], [23]–[27].

A typical processing pipeline for supervised detection algorithms begins with a "prescreening" operation, in which a relatively fast algorithm is applied to the full set of GPR data (e.g., a 3D volume, or B-scan) in order to identify a smaller subset of spatial locations on the ground, which are subsequently processed by the supervised algorithms. Prescreening is used primarily because it dramatically reduces the amount of data required for both training and testing supervised algorithms, making it possible to apply such algorithms in real-time applications (e.g., on a truck while driving). In the second step of processing, one or more 2D patches of GPR data are extracted at each prescreener alarm location, and a (trained) supervised algorithm is applied to the patches in order to predict whether each alarm location is a threat, or non-threat.

In order to train supervised classifiers, they must be provided with examples of data from each class (i.e., threat and non-threat). As mentioned, training examples most often consist of small image patches that are extracted from B-scans at locations in the GPR volume where useful signals (i.e., those corresponding to both threats, or suspicious non-threats) are estimated to exist [1], [4], [6]–[8], [12], [28]–[32]. In this work, we refer to these useful signal locations as "keypoints". The performance of supervised algorithms depends strongly upon the training data that the algorithm is provided, and as a result, the way in which keypoints are (i) identified and (ii) utilized, forms an important design choice for supervised GPR algorithms.

*A. Keypoint identification*

A GPR keypoint consists of a spatial location as well as a temporal location (or sometimes depth). The way in which keypoints are identified is fairly consistent in most existing GPR-based threat detection research. The spatial location is typically either (i) known in advance because the objects were deliberately buried [1], [6] or (ii) it is estimated using a detection algorithm (sometimes called a prescreener) that precedes supervised classification algorithms [1], [6], [7], [11], [12], [33], [34].

Once the spatial location is obtained, the temporal location can be estimated. By far, the most common approach for temporal estimation relies on extracting keypoints at locations of high energy (e.g., local maxima) in the GPR A-scans [6], [8], [20], [29], [35]–[39]. These energy-based methods often yield multiple keypoints at each spatial location. Figure 2 illustrates a previously proposed method [37], which we term the max-smoothed-energy keypoint (MSEK) approach. MSEK is used in this work, and is representative of most existing temporal identification approaches, though some others do exist [8], [10], [39]–[41].

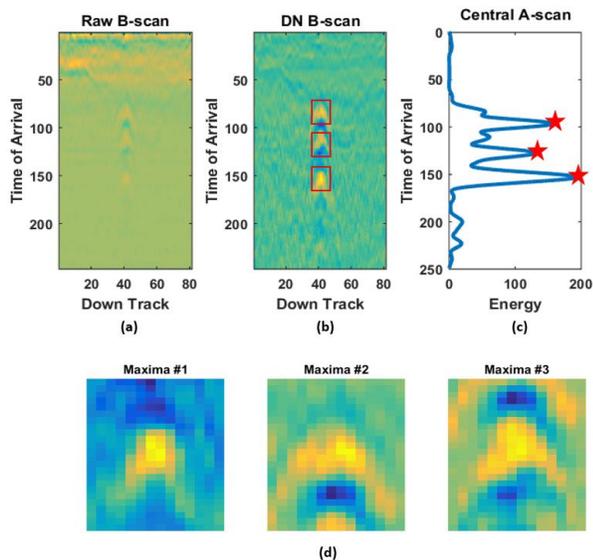

**Figure 2:** the process of identifying the temporal location of GPR keypoints with MSEK. In MSEK, (a) the raw data is (b) depth normalized, (c) the central A-scan is squared and smoothed and keypoints are identified by the maximum values in the transformed, central A-scan. At the top maxima locations, indicated by stars in (c), patches are extracted, where the three examples shown in (d) correspond to the data in the red boxes drawn in (b).

*B. Keypoint utilization*

Although there is a general consensus in the literature about how to identify keypoints (i.e. that regions of high radar amplitude are of interest), there is relatively little agreement regarding keypoint utilization. Here, keypoint utilization will refer to the process of deciding which keypoints, of those identified, that should be provided to the supervised classifiers for training, as well as testing.

Consider first the problem of keypoint utilization for training. Keypoint identification approaches, such as MSEK, yield multiple keypoints at each spatial location. As a result, it is unclear which keypoints should be retained for training, and this ambiguity is evidenced by the large number of keypoint utilization approaches that have been proposed in the literature (see section II). To date, it is unclear which of these utilization strategies are best, or whether there are any generally superior approaches at all.

Similar design choices must be made for keypoint identification during algorithm testing as well: given a spatial location generated by a prescreener, which temporal locations should be evaluated by the (trained) classifier? Further, given multiple decision statistics at each spatial location, how should a final, single, statistic be computed? Similar to utilization for training, a large variety of utilization approaches have been employed in the literature during testing [9], [18], [34], [35], [38], [41], [44], and it is likewise unclear which existing strategies, if any, are superior.

*C. Contributions of this work*

In this work we investigate the problem of keypoint utilization during both training and testing. We present a taxonomy of existing keypoint utilization strategies, and then compare their effectiveness on a large dataset of GPR data using several combinations of state-of-the-art classifiers and features. The results indicate that keypoint utilization can have a significant impact on the classification performance, and that some utilization strategies do generally outperform others. Further, we introduce and apply a new utilization method that was inspired by the comparison of existing methods. This proposed method, called "PatchSelect", outperforms all existing methods across all combinations of features and classifiers tested in this work. As further analysis, we present several small-scale experiments that motivated the design of PatchSelect, as well as elucidating effective practices for keypoint utilization. In the conclusions section, we also discuss limitations of our work, including PatchSelect.

The remainder of this paper is organized as follows. In Section II, we present a taxonomy for the different keypoint utilization methods that have been proposed in the literature. In Section III, the experimental design of a large-scale comparison of keypoint utilization strategies is explained. In Section IV, the results of the comparison are presented. In Section V, the small-scale experiments motivating PatchSelect are presented. In Section VI, conclusions are drawn and recommendations are given for the design of future BTD systems.

## II. KEYPOINT UTILIZATION STRATEGIES

In this section, we present a taxonomy of existing keypoint utilization strategies that have been employed in the literature. Keypoint utilization strategies, as defined here, consist of two components: one component for training, and one component for testing. We will present a taxonomy of existing methods based on a few characteristic differences they have during both training, and testing. Table 1 in this section presents the taxonomy of each existing method, as well as our proposed method, PatchSelect.

### A. Strategies for training

During training, there are two main characteristics that differentiate existing keypoint utilization strategies: the number of keypoints extracted at each spatial location, denoted by $K$, and how this number varies between target locations and non-target locations.

For example, many strategies utilize keypoints at the top $K$ energy locations, where $K$ is the same for both target and non-target cases [2], [6], [9], [18], [20], [34], [37]–[40]. In contrast, in [29], [42], [43], a different $K$ is used for target and non-target cases. Other strategies use energy keypoints for target cases but extract data for non-target cases at regular or random intervals down the A-scan (i.e., no estimation of non-target data localization is performed) [7], [19], [27], [42], [43], [44]. Finally, in [27], every time point is considered in training as a keypoint for both classes. These methods are listed in Table 1 where each method's keypoint utilization strategy for target and non-target is listed.

### B. Strategies for testing

The characteristics that differentiate utilization strategies during testing are very different than those during training. The primary reason for this is that, during testing, the primary goal of utilization is to obtain a single decision statistic for each *spatial* location. Every spatial location consists of many potential keypoints (one keypoint at each temporal location), and therefore we must decide: (1) how many total keypoints, denoted by $L$, should be utilized to compute a final decision statistic at that location; and (2) upon what criteria (e.g., energy) we should choose these keypoints. In the remainder of this section we explain the common approaches existing utilization methods take to address these questions, and we include these into our taxonomy in Table 1.

Existing strategies typically choose which keypoints to utilize based on one of two ordering criteria. The first ordering criterion is to utilize keypoints at maximum energy locations, in the same way it is done during training [8], [10], [35], [39], [45]. This is denoted as "En" in Table 1. The second ordering criterion is to utilize the largest classifier decision statistics, denoted as "DS" in Table 1. In this scenario, the classifier is applied at regular intervals along the A-scan and keypoints at the $L$ locations with the largest classifier decision statistics are utilized [6], [7], [11], [28]. The strategies in [8], [10], [20], [35], [36], [39], [45] set $L = K$, so that the same number of keypoints are utilized in training and testing.

Once an ordering criterion for the set of $L$ testing keypoints is chosen, a final decision statistic must be computed. At a spatial location, the strategy's ordering criterion (i.e., En/DS) is used to organize the decision statistics in to a decreasing sequence $D = \{d(j); j = 1 \ldots L\}$ from which a final decision statistic $D_f$ is computed. Several approaches are taken toward obtaining a final decision statistic. First, if En is the ordering criterion, then $D_f = \max D$ for all such methods listed in Table 1 [2], [5], [8], [39]. Second, if DS is the ordering criterion, a function $g$ is typically defined to compute $D_f$. Several different options for $g$ have been considered such as: max, sum over all $T$ (all time points), or mean of the top 3 decision statistics [9], [18], [34], [6], [16], [27], [29], [43], [46], [47]. Note that these three options for $g$ can be parametrized. If $g$ is defined by

$$g(D, L) = \sum_{j=1}^{L} d(j) \quad (1)$$

then for $L = 1$, $g(D, 1) = d_1 = \max(D)$. Similarly, if $L = T$, then $g(D, T) = \sum_{t=1}^{T} d(t) = \text{sum}(D)$. Finally, if $L = 3$, then $g(D, 3) = d(1) + d(2) + d(3)$. This is not the same numeric value as the average of the top 3 confidences, but every output of $g$ is a constant scaling factor from the average (i.e., off by a factor of 3) and such a factor does not affect the *probability* of detection and of false alarm. Because of the similarity of the three different methods employed in the literature, they are listed in their parametrized form in Table 1.

Table 1: A table of existing keypoint utilization strategies, including the proposed method, PatchSelect. In training, the number of patches for targets and non-targets are given. In both cases, if no parenthesis is written, then the locations are chosen using an energy-based temporal localization method (e.g., MSEK). The demarcation of "reg" means that that number of patches were extracted at regular intervals, and "rand" means that they were extracted at random. In testing, the number $L$ specifies the number of decision statistics used to obtain a final confidence. The column "En/DS" denotes whether the locations from which the decision statistic is taken depends on its (top $L$) maximum energy locations, "En", or whether the top $L$ decision statistics, "DS", are used. Method 9 is unique in that a sliding window operation is first performed on the decision statistics where the 7 consecutive confidences are summed.

| | | Train | | Testing | |
|---|---|---|---|---|---|
| Index | Reference | #target | #non-target | En/DS | L |
| 1 | [6] | 3 | 3 | DS | 3 |
| 2 | [8], [39] | 2 | 2 | En | 2 |
| 3 | [42]–[44] | 1 | 5 (reg) | DS | 3 |
| 4 | [29] | 1 | 5 (rand) | DS | T |
| 5 | [36] | 1 | 1 | DS | 12 |
| 6 | [20] | 1 | 1 | DS | 1 |
| 7 | [21] | 5 (reg) | 5 (rand) | DS | T |
| 8 | [5] | 3 | 3 | En | 1 |
| 9 | [11] | 1 | 1 | DS | sliding max |
| 10 | [2] | 1 | 1 | En | 1 |
| 11 | PatchSelect | 4 | reg | DS | 12 |

### C. PatchSelect

The training and testing strategies described in this section are summarized in Table 1, along with the proposed PatchSelect method. PatchSelect consists of training on the top 4 energy keypoints for targets and on patches extracted at small, regular intervals along the central A-scan for non-targets. In testing, the sum of the top 12 decision statistics is reported as the final confidence.

To determine the specific design choices for the PatchSelect strategy, a series of experiments were conducted (described in

Section V) to identify which utilization characteristics tend to yield better performance. In particular, we investigate the impact of several of the characteristics in Table 1 that we used to taxonomize existing methods (e.g., how many $H_0s$ to use in training). The results of these experiments reveal good general practices for keypoint utilization, and motivated our design of PatchSelect.

## III. EXPERIMENTAL SETUP

In this section we present an experimental method used for comparing the keypoint utilization strategies shown in Table 1. The utilization strategies are compared by evaluating the performance of several state-of-the-art feature sets and classifiers when each of the keypoint utilization strategies are employed during training and testing.

### A. Evaluation dataset

The data used in this work was collected using a downward looking GPR (similar to the one described in [33]), at a western U.S. test site over a total area of 167,167.3 m². The data was collected over 8 test lanes for a total of 75 runs over all of the lanes. The dataset includes 1,967 target encounters. From this collection of data, spatial and temporal locations must be identified to act as training and testing data for targets and non-targets.

To identify spatial locations of interest for targets and non-targets, a prescreener can be used [1], [8], [27], [32], [43]. To correspond alarms with targets, GPS data is collected along with the GPR data, where the GPS locations of buried threats is known *a priori*. The remaining alarms correspond to non-targets and are used for training the algorithm to recognize non-targets, because those are the instances it will have to classify at test time. In this work, the energy-based F1V4 prescreener [48] is used, which identifies locations with anomalous energy profile compared to the relatively unchanging background. A sensitivity threshold is set for the prescreener which yields a dataset of 1,771 target alarms and 640 non-target alarms. This threshold was chosen to achieve the highest possible probability of detection with this prescreener at an operationally feasible false alarm rate (number of false alarms per unit area).

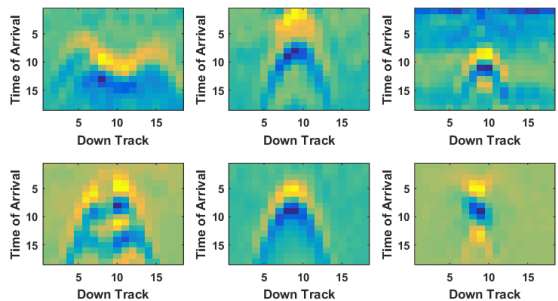

**Figure 3: Examples of extracted patches at keypoint locations at known target locations. The patches capture some part of the target signature which is used during training as an example of general target signature appearance.**

### B. MSEK for keypoint identification

Given spatial locations generated by the F1V4 prescreener, keypoint identification was performed using the MSEK algorithm which was first introduced in [37] and has been employed in many studies [6], [19], [21], [36], [49]. The process of obtaining temporal keypoints using MSEK is illustrated in Figure 2. In MSEK, the data is depth normalized, the central A-scan is squared and smoothed and keypoints are identified at the maximum values in the transformed A-scan. A patch of data can then be extracted surrounding at each maximum location, examples of which are shown in Figure 3. While other methods exist for temporal keypoint localization, MSEK is representative of methods relying on the measured amplitudes, and is simple to implement and use. For this reason, only MSEK is used for keypoint identification with this data set.

### C. Feature sets and classifiers

The different keypoint utilization approaches that are summarized in Table 1 are evaluated using several combinations of features and classifiers to investigate which methods tend to be the most effective. Following the approach for evaluating a BTD algorithm described in [6], features are extracted at each keypoint and are used to train and test classifiers. These features and classifiers were chosen because they have been used frequently in the GPR buried threat detection literature [6], [8], [19], [28], [38]. The resulting feature and classifier combinations are referred to as BTD algorithms.

In this study, we consider several sets of features that have recently been applied for BTD with GPR: the raw data (rasterized) (e.g., [24], [50]), histogram of oriented gradients (HOG) features (e.g. [6], [8], [51]), and edge histogram descriptor (EHD) features (e.g., [7], [28], [52]). The data patches on which features are extracted are of size $18 \times 18$ pixels and rescaled to have values between $-1$ and $1$. To match the results from [6], the parameter choices for HOG and EHD are kept the same: the HOG feature is computed in cells of $6 \times 6$ pixels, normalized in blocks of $3 \times 3$ cells, and 9 angle bins and the EHD feature is computed with a threshold of 0.15. The threshold for declaring a gradient as "no-edge" in EHD depends on the scaling of the data.

We used two classifiers in this work: a radial basis function SVM [53] and a random forest (RF) classifier (100 trees, 2 variable splits at nodes, with central axis projection) [54]. These two classifiers were chosen because of their recent application to GPR resulting in state-of-the-art detection performance [6], [8], [12], [36], [45], [55], [56].

### D. Cross-Validation and performance metrics

In this work we trained and tested each classifier using four-fold cross-validation. This is a common approach for evaluating the performance of machine learning algorithms, and has been employed previously for BTD with GPR data [6], [20], [28]. Additional care had to be taken in our experiments because the GPR data is collected over the same lane multiple times. As a result, all alarms within a certain spatial distance were clustered and assigned to the same fold, to avoid training and testing over the same physical area. To properly handle the issues associated with proper cross-validation on this type of data set, researchers at the University of Florida developed software which is used here and has been used in many previous studies [1], [6], [7], [11], [29], [34].

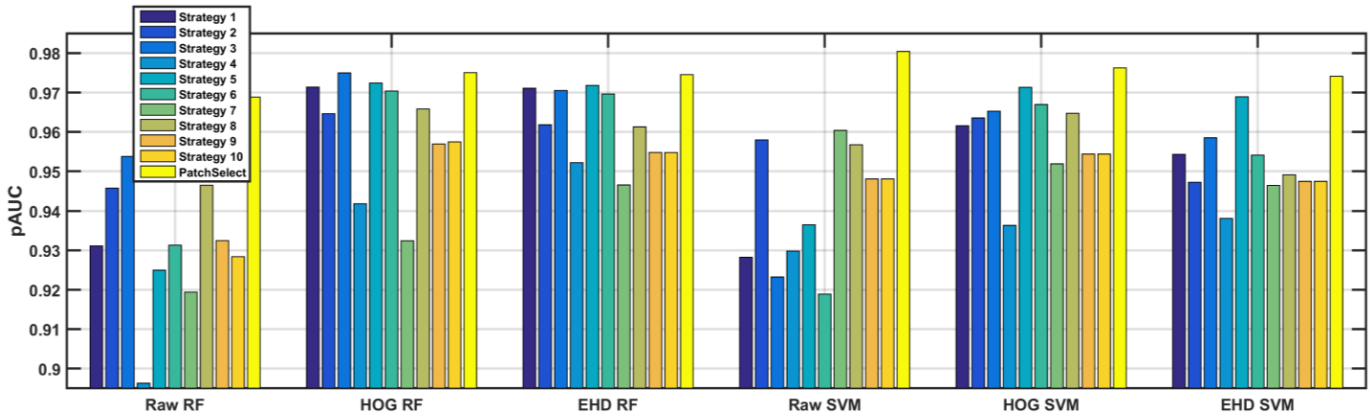

Figure 4: pAUC of the 11 utilization strategies listed in Table 1 are shown with pAUC calculated up to a FAR of 0.005. Each group of 11 bars represents a separate feature and classifier combination (listed on the x axis) where the random forest is denoted as "RF". The 11 bars can be compared for their effectiveness within group and bars of the same color, representing the same strategy, can be compared across condition.

To compare the detection performance of each trained classifier, receiver operating characteristic (ROC) curves are used. ROC curves are a common metric for comparing machine learning algorithms, and they are likewise popular in the BTD algorithm research literature [1], [6], [7], [11], [29], [34]. ROC curves plot the relationship between the false detection rate (x-axis) and true detection rate (y-axis) of a detection algorithm, as the sensitivity of the algorithm is varied. In the BTD literature, it is common to scale the x-axis of the ROC curve to report the false alarm rate in terms of false alarms per square meter [1], [11], [56], and we adopt this practice here.

The ROC curve can also be summarized with a single statistic. One commonly-used statistic for this purpose is the partial area under the ROC curve (pAUC). This metric is obtained by computing the area under the ROC curve between two false alarm rate (FAR) values (e.g., 0 and 0.005 FAR). pAUC is frequently used for performance comparisons in the BTD literature [33], [37], [55], [57], [58].

IV. COMPARISON OF KEYPOINT UTILIZATION STRATEGIES

This section presents the results from the comparison of the eleven keypoint utilization strategies listed in Table 1 and discussed above.

A. *Performance of keypoint utilization strategies*

Results of this evaluation are shown in Figure 4 where the pAUC of the 10 strategies from the literature and PatchSelect are grouped by their performance under the different feature and classifier combinations. We make several observations of the outcome indicated by these results.

First, the choice of keypoint utilization strategy can have a large impact on performance. For example, choosing PatchSelect over strategy 3 for the Raw SVM feature and classifier combination yields a pAUC improvement of 0.058. Furthermore, if strategy 6 is used for keypoint utilization, the Raw SVM BTD algorithm would be considered among the worst performers among the 6 algorithms, whereas with PatchSelect it is the best performer across all conditions. The variance in performance for a single feature and classifier combination suggests that choosing a poor performing keypoint utilization strategy could negatively bias the results of a new BTD algorithm that may have merit when accounting for the training and testing variance.

Second, certain strategies are consistently among the top performers (e.g., strategies 3, and 5). The average rank of top performing strategies such as 3 and 5 across the 6 BTD algorithms is 3 and 3.7 respectively. This suggests that certain practices are generally good for keypoint utilization when training and testing BTD algorithms. This result is important because it implies that by using those good practices as a strategy for keypoint utilization, the possible loss in performance will be due to the algorithm design and not due to how it is trained. Thus, using identified best practices simplifies evaluating the performance of BTD algorithms.

Third, Figure 4 shows that PatchSelect outperforms all other strategies for the 6 BTD algorithms except for HOG RF where strategy 3 and PatchSelect are tied. We developed PatchSelect by identifying best practices among other existing strategies and incorporating them into a single training and testing strategy. The practices suggested by PatchSelect seem to provide generally stable results even if the approach is changed slightly, as described in section V. The limitations of this comparison and these conditions are discussed in the conclusions section.

Fourth, some strategies in the above comparison seem to consistently perform at the bottom. A caveat about these methods (in particular [21], [29], strategies 7 and 4 respectively) is that they were designed for a different classification paradigm than the one used in this work, namely, Multiple Instance Learning.

B. *Performance sensitivity to varying pAUC measures*

The pAUC measures presented in Figure 3 are computed over a specific range of FAR values (i.e., 0 to 0.005 FAR). This metric summarizes the performance of each classification algorithm only over the aforementioned FAR range, and therefore the results may not hold for a different FAR range. In this section we evaluate the pAUC of the classification models as we vary the FAR range in the pAUC computation. In particular, we vary the larger of the two FAR values, which we term $FAR_2$, and compute the average pAUC of each keypoint utilization approach. These results are shown in Figure 5.

The results of this analysis indicate that the rank order of the different strategies remains relatively unchanged as the FAR range is varied. Further, for each FAR value, the pAUC of PatchSelect (strategy #11) is higher than that of any other strategy.

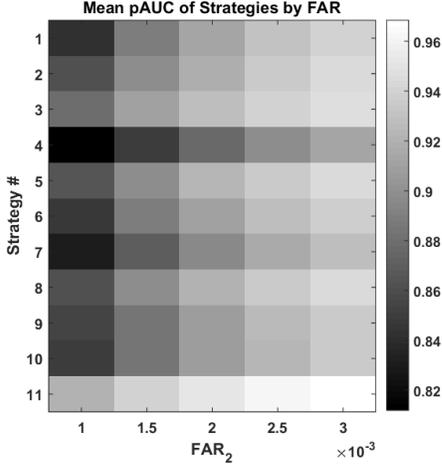

Figure 5: Average pAUC across the 6 conditions of features and classifiers of each strategy is computed in each column at an increasing maximum FAR value.

## V. MOTIVATION FOR THE PATCHSELECT STRATEGY

To understand the specific parameter choices for PatchSelect given in Table 1, a study of the design choices of existing methods is conducted in this section. As discussed in section II, each strategy defines a particular way to train (e.g., use 3 keypoints at maximum energy locations for both classes) and a particular way to test (e.g., sum the top 3 decision statistics). The experiments in this section address the sensitivity in performance associated with these choices. For brevity, the results of these questions are shown here using the HOG feature with the random forest classifier (denoted HOG-RF). This choice is somewhat arbitrary, but it was chosen because trends exhibited by HOG-RF are fairly consistent across other features and classifiers. Additionally, PatchSelect was shown in Figure 4 to outperform other methods consistently across all tested BTD algorithms, suggesting that the conclusions drawn from HOG-RF are indeed general. Further, the comparison of performance as a function of the final FAR threshold in section IV.B suggested that the rank ordering of the results is relatively insensitive to that threshold. For this reason, the results of these experiments are shown for a pAUC computed to a FAR of 0.005.

### A. How keypoints should be chosen at test time: energy or classifier confidence?

Strategies differ in their use of keypoints at test time, between energy and classifier decision statistics. In this section we try to examine whether one of these two approaches tends to be superior to the other. To do this, we compared these two general approaches while controlling for many other experimental factors (e.g., different training strategies for targets and non-targets). The results of this comparison are presented in Figure 6, which suggest that using the top decision statistics outperforms using decision statistics at maximum energy locations.

### B. How many testing keypoints should be used?

In each strategy, the number of testing keypoints, $L$, has to be specified. In this section, we examine how to determine that number. This number depends on whether energy maxima, or decision statistics, are needed. The results presented in Figure 6 suggest that using $L = 4$ for energy maxima works best whereas, $6 \leq L \leq 12$ is more suitable when decision statistics are used.

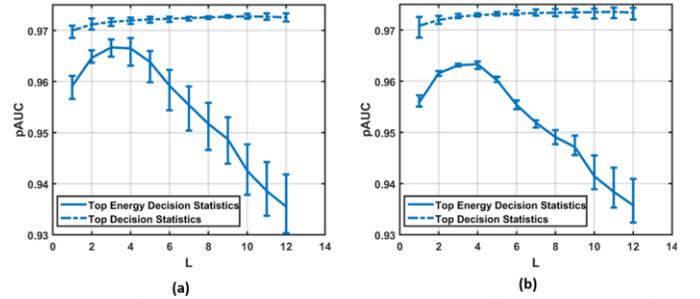

Figure 6: Performance comparison between two different strategies for obtaining a final confidence: using the top $L$ decisions (dashed line) or using the $L$ decisions at top energy-locations (solid line). The 2 subplots refer to two different non-target training strategies: (a) training on the top $K$ energy locations (b) at 5 regularly extracted patches. In both plots, each curve represents the average performance when varying then number of target patches used in training. The error bars show the range of performance obtained across the different target training strategies (i.e., training with 1-4 patches for targets).

### C. How should non-target training keypoints be chosen?

While all existing utilization strategies train on data from maximum energy locations for targets, this is not the case for non-targets. The methods compared in Table 1 utilize keypoints extracted at the top $K$ locations (the same $K$ is used for target data) or at 5 regularly spaced indexes. In this section, we examine the choices for providing non-target training data to a classifier.

We note that there are 2 main differences between the two proposed $H_0$ approaches. The first is that extracting data at energy locations is a physics-based criterion which may be superior to extracting data at regular intervals (this may include data at regions outside non-target signatures and thereby negatively biasing the classifier). The second is that the approaches that extract data at regular intervals do so with more patches (5) than the energy based methods (the most is 4 patches). Thus, performance improvements may be solely on the basis that more patches are being provided to the classifier when using the strategy of extracting data at regular intervals. For this reason, a third training condition is added where non-target patches are taken down the depth at every fourth location down the A-scan (82 patches per non-target observation in total) which corresponds to approximately 75% overlap between consecutive patches. This condition is added to test whether more data would improve classification performance.

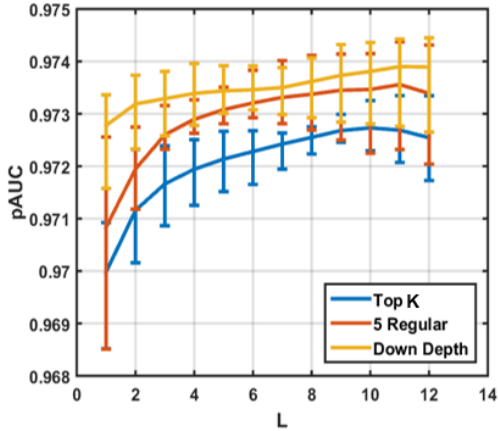

**Figure 7:** Non-target training strategies are compared using the confidences obtained using the top $K$ decision statistics. The error bars represent the variance in training on 1-4 patches per target alarm and the plotted line represents the average performance.

To present this comparison, the final confidence is computed on the top decision statistics, as this was shown in section V.A to be superior. In Figure 7, the three methods for choosing training data representing non-target data are compared. During testing, the final confidence is obtained by summing the top $L$ decision statistics. For all values of $L$, the strategy of providing more non-target data to the classifier improves performance (i.e., "Down Depth" method).

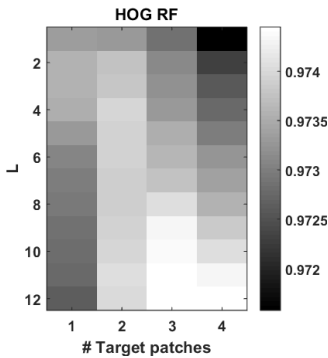

**Figure 8:** Performance when training on 1-4 patches per target alarm. This comparison is performed when training on non-target patches extracted at small, regular intervals and the final confidence relies on the top $L$ confidence locations.

### D. How should target training keypoints be chosen?

The final design question we considered regards the number of target patches that should be used in training. In this section, we address this question in the context of the answers obtained in previous sections. Therefore we train on non-target data extracted at small regular intervals; and during testing, we summed the top $L$ confidences. The results suggest that the best performance is achieved using $K = 4$ for target data.

The results of these experiments are shown in Figure 8. The results indicate that training with 4 patches, extracted at energy maxima locations, performs best. As was explained in section III.B, MSEK chooses local maxima and in our data, the temporal extent of target signatures does not typically extend beyond having 4 local energy maxima. This suggests that training on patches with some portion of the target signature is beneficial for the classifier.

## VI. Conclusions and utilization recommendations

In this work, the question of how to choose training and testing data for supervised GPR BTD algorithms is addressed. Training and testing data consist of small patches of GPR imagery, which we refer to here as "keypoints". While most algorithms in the GPR literature identify keypoints in a very similar fashion, there is much variability in how they are utilized once they are identified. In this context, utilization refers to several design questions: choosing, among identified keypoints, which keypoints should be provided to supervised classifiers (during training); to which keypoints the classifier should be applied during testing; and how a final decision statistic (or confidence) should be computed using the keypoints. A large variety of methods have been proposed in the literature for this purpose, and it is unclear which approaches are best, or whether any methods are superior to others.

In order to address these questions, we compared the effectiveness of many existing keypoint utilization approaches on a large GPR dataset, and using a variety of different classifiers and features from the GPR literature. We also proposed a new method, called PatchSelect, which was designed based on insights from our experiments in this work. Based on the results, several conclusions can be drawn:

- The choice of utilization strategy has a significant impact on the detection performance of the resulting supervised algorithm.
- There are utilization practices that generally yield better results.
- We combined the best identified practices to create the PatchSelect strategy, which (in our experiments) is always superior.

In addition to our large-scale comparison, we also conducted several smaller experiments (Section V) to elucidate which utilization practices yield the best results. These experiments also motivated the design of PatchSelect. In the last subsection here, we make a few additional recommendations and comments for keypoint utilization, based on the results of our smaller experiments Section V.

### A. General recommendations for keypoint utilization and applying PatchSelect

The goal of this investigation was to identify best practices during training and testing of GPR BTD systems. This was identified as an important problem because the comparison in section IV.A shows the possible variance of a single method across conditions and the change in rank ordering between methods across conditions. PatchSelect represents the set of practices that were found to be best on a large collection under several conditions. There are, however, some limitations which are addressed in this section.

The first limitation is the possible computational burden from training with the PatchSelect training dataset. PatchSelect entails training on the top 4 energy locations for targets and training down the depth for non-targets. The results in section V.C suggest that training on non-target patches with a bit less than 75% overlap would not degrade performance too much. Similarly, the results in section V.D suggest that the difference between choosing 3 or 4 target patches is very slight. In both cases, the difference in pAUCs is < 0.007 for the 6 features and classifier combinations tested here. The most important factor seems to be to sum the top decision statistics during testing. However, the results suggest that if more training data is needed for a particular algorithm, those additional patches are suitable.

A second limitation of this work is its use of MSEK as a temporal localization method. In [49], the energy related to target signatures was shown to be relatively localized, which suggests that data only around the energy maximum should be used in training and testing. However, the performance of this approach is consistently less than summing the top $L$ decision statistics. Experimentally, we note that summing decision statistics rather than the decisions at energy maxima works best when the data from training is representative of the data that observed at test time. In this context, this means to train on data down the depth if testing is done down the depth. Balancing the dataset in this way is an important principle in designing training and testing sets [59]. This result motivates an investigation into MSEK whose locations may be inconsistent between training and testing. In this work, because the sum of the top decision statistics is more stable and generally performs best, it is recommended.

A third limitation is the dataset which was used in this work. Although six different feature and classifier combinations were compared and their pAUC is computed at several FAR values, the alarms in the dataset remain the same. While trends exist in the presented results, the conclusions may depend upon the specific choice of keypoint identification methods used in this work.

## ACKNOWLEDGEMENTS

This work was supported by the U.S. Army RDECOM CERDEC Night Vision and Electronic Sensors Directorate, via a Grant Administered by the Army Research Office under Grant W911NF-06-1-0357 and Grant W911NF-13-1-0065.